# Building Change Detection for Remote Sensing Images Using a Dual Task Constrained Deep Siamese Convolutional Network Model

*Yi Liu, Chao Pang, Zongqian Zhan, Xiaomeng Zhang, and Xue Yang*

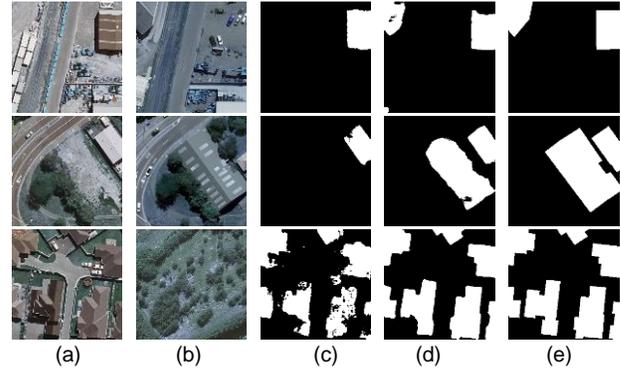

Fig. 1. Example results of DTCDSCN and improved SegNet on the WHU building dataset. (a) 2016 images. (b) 2012 images. (c) Result of improved SegNet. (d) Result of DTCDSCN. (e) Ground truth.

*Abstract*—In recent years, building change detection methods have made great progress by introducing deep learning, but they still suffer from the problem of the extracted features not being discriminative enough, resulting in incomplete regions and irregular boundaries. To tackle this problem, we propose a dual task constrained deep Siamese convolutional network (DTCDSCN) model, which contains three sub-networks: a change detection network and two semantic segmentation networks. DTCDSCN can accomplish both change detection and semantic segmentation at the same time, which can help to learn more discriminative object-level features and obtain a complete change detection map. Furthermore, we introduce a dual attention module (DAM) to exploit the interdependencies between channels and spatial positions, which improves the feature representation. We also improve the focal loss function to suppress the sample imbalance problem. The experimental results obtained with the WHU building dataset show that the proposed method is effective for building change detection and achieves a state-of-the-art performance in terms of four metrics: precision, recall, F1-score, and intersection over union.

## I. INTRODUCTION

In the field of remote sensing, the detection of change is a process that involves utilizing remote sensing images acquired at different times to identify the changes that have occurred over the same area. Building change detection is extremely important in the fields of land-use planning, city management, and emergency response. However, manual change detection is time-consuming and labor-intensive, so there is a need for automatic and efficient change detection.

Building change detection algorithms can be summarized into two categories: 1) post-classification comparison based

Manuscript received XXXX, 2018; revised XXXX; accepted XXXX. Date of publication XXXX; date of current version XXXX. This work was supported in part by the National Key Research and Development Program of China under Grant 2016YFC0802500, in part by the China Space Foundation under Grant 6141B06240203, and in part by the National Natural Science Foundation of China under Grant No. 61871295. *(Corresponding author: Zongqian Zhan.)*

The authors are with the School of Geodesy and Geomatics, Wuhan University, Wuhan 430079, China (e-mail: zqzhan@sgg.whu.edu.cn).

Color versions of one or more of the figures in this letter are available online at http://ieeexplore.ieee.org

Digital Object Identifier XXXXXX

methods; and 2) direct classification based methods. Post-classification comparison [1]–[2] involves extracting the buildings from the images of different phases, and then obtaining the change detection classification result by comparing the extracted building maps. There is therefore a need for a high accuracy of building extraction for both images, and there is a serious possibility of accumulated error. The general process of direct classification, which requires only one classification stage, is to first generate a similar feature map, and then analyze the map to achieve a change map. A building change detection method based on the morphological building index was proposed by Huang *et al.* [3], which combines the morphological index and spectral features and shapes as the change conditions, and extracts the building change regions by setting thresholds. However, this feature extraction method cannot be widely applied because it is complicated and requires many empirical thresholds.

With the development of deep learning, some scholars have turned to the study of change detection based on deep learning. Zhan *et al.* [4] used a deep Siamese convolutional network to extract similar feature maps, and then clustered them with a *k*-nearest neighbor (KNN) classifier to obtain the changed area. Similarly, Liu *et al.* [5] used a symmetric convolutional neural network to extract multi-temporal image features, and then transformed them into features in a consistent space, which also achieved good results. These approaches based on deep learning are capable of extracting more robust features and have demonstrated good performances. However, the methods mentioned above are all done step by step, which is not ideal. The introduction of the fully convolutional neural network has led to great breakthroughs in the field of semantic segmentation. Inspired by this, Zhu *et al.* [6] introduced and improved the SegNet neural network for building change detection, which is a method that can be end-to-end trained.

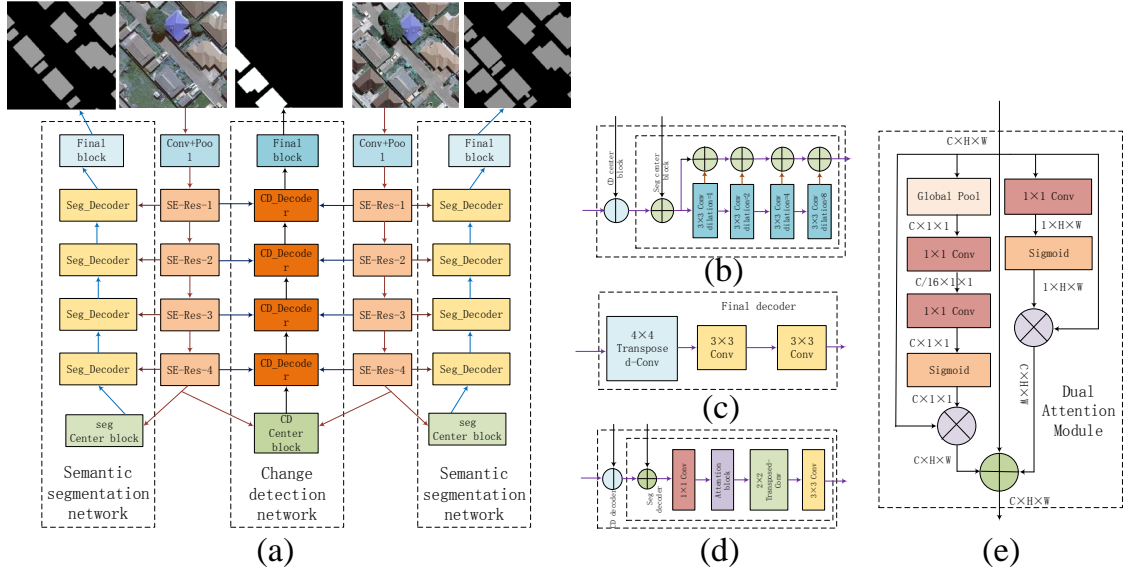

Fig. 2. An overview of the dual task constrained deep Siamese convolutional network. (a) Network architecture. (b) Components of the center block. (c) Components of the final block. (d) Components of the decoder block. (d) Components of the dual attention module.

This method can efficiently obtain a pixel-level building change map, but it does have some shortcomings: 1) it is insensitive to changes in buildings with small spectral and structural differences, as shown in the first and second rows of Fig. 1; and 2) the extracted changed areas often have holes and their boundaries can be very irregular, resulting in a poor visual effect, as shown in the third row of Fig. 1.

The above issues are the main reason that the learned features are not discriminative enough, which is also the main challenge of change detection. To tackle the challenge, we address the causes of the problem from the perspective of the second shortcoming. We can see that the building features in the bitemporal images of the third row of Fig. 1(c) are obvious, but the obtained building change regions are incomplete. However, this situation is greatly improved with the semantic segmentation since improved SegNet tends to learn the features of pixel-level differences in bitemporal images. As a result, the object-level features extracted by the proposed model contribute to more precise extraction results, and are more discriminative for the building change detection task. To this end, in this paper, we propose a dual task constrained deep Siamese convolutional network (DTCDSCN) model to jointly optimize the building change detection and semantic segmentation tasks. In summary, this letter makes three main contributions:

1) We propose a novel building change detection network model, which can simultaneously carry out the main task of building change detection and the auxiliary task of building extraction. The two tasks share the same feature extraction layer, so the auxiliary task enables the model to learn discriminative object-level features, which contribute to more precise building change detection results.

2) We introduce a dual attention module (DAM) to further improve the discriminative ability of the features, which can take advantage of the interdependencies between channels and spatial positions.

3) We propose the change detection loss (CDL) function by improving the focal loss (FL) [7], to address the sample imbalance problem in building change detection. The relationship between the weights of the changed and unchanged samples is closer to reality in CDL.

The rest of this letter is organized as follows. The proposed method is described in Section II. The experimental results and a discussion are presented in Section III. Finally, the conclusion of this paper is drawn in Section IV.

## II. PROPOSED METHOD

The DTCDSCN model we propose consists of a change detection network and two semantic segmentation networks (SSNs), as shown in Fig. 2(a). By simultaneously performing the change detection and semantic segmentation tasks in a unified framework, the proposed model can focus on the object-level features of buildings, and can thus improve the discriminative performance of the features. In terms of the change detection task, the number of changed samples is smaller than that of unchanged samples, which results in extreme imbalance of the sample sets. Consequently, we developed the CDL function on the basis of the FL function.

### A. Siamese Change Detection Network

The proposed change detection network structure is based on the Siamese network structure, which is in the form of a fully convolutional neural network, while the overall design follows the encoder and decoder style. In detail, the network structure contains two encoding branches, both of which share trainable weights, and one decoding branch. The two encoders share weight values, which are used for the feature extraction of the images at phase 1 and phase 2, respectively.

We take the SE-ResNet [8] network structure as the basic encoding module of our proposed network structure, which involves five stages in total. In order to take advantage of global context information, we introduce the spatial pyramid pooling module as the center block, as in [9], to enlarge the receptive field of the feature maps and to embed different-scale contextual features.

Symmetric with the encoding part, the decoding part also contains five parts, and we use the D-LinkNet [9] decoder as the baseline model. In detail, the first four decoders are the same, as shown in Fig. 2(d), where each change detection block (CD block) contains three inputs. Considering that features from different spatial positions and different channels may be correlated, and inspired by the attention mechanism, we add the DAM into the decoder part, which takes both the spatial and channel attention mechanisms into consideration, so that we can improve the discriminative ability of the feature extractor. The fifth decoder is the final block, as shown in Fig. 2(c).

The structure of the DAM is illustrated in Fig. 2(e). The DAM obtains the channel-level attention feature map and the spatial-level attention feature map of the input feature map, and we then combine the channel-level attention feature map with the spatial-level attention feature map to obtain the final output feature map. The attention feature map can make full use of the interdependencies within channels and the spatial positions, and is combined with the input feature map by a shortcut connection operation, which can improve the discriminative ability of the features, without destroying the information.

### B. Semantic Segmentation Network

Since the change detection and semantic segmentation tasks are similar, the SSN we use is similar to the change detection network. However, the SSN alone can only be applied to images of a single time phase.

### C. Network Architecture

We propose the DTCDSCN model to integrate the change detection task with the semantic segmentation task. Through learning the semantic segmentation of buildings, it provides guidance for the model to learn the object-level features of the buildings, which can make the extracted features more discriminative and obtain better change detection results. The proposed model is based on the Siamese architecture, and has two encoders that share weight values to extract the feature maps of the bitemporal images. Apart from the decoder part of the main change detection network, our dual-task network also has two SSNs sharing weight values. The model receives bitemporal images, and outputs a change detection map and the segmentation results of the bitemporal images. We use deep supervision to obtain better results and make the network easier to optimize. In the application of change detection, the number of changed samples is much smaller than that of unchanged samples. In terms of the sample imbalance, the FL function proposed by Lin *et al.* [7] is the most effective solution, so we developed the CDL function by improving the FL function. Therefore, for the change detection network, we use the CDL function. For both SSNs, we use a cross-entropy loss function. Finally, we use two weight parameters to balance the CDL and the two segmentation loss functions in the loss function of DTCDSCN, as shown in (3).

$$\ell_{ss} = BinaryCrossEntropy(\hat{y}; y) \tag{1}$$

$$\ell_{cd} = CDL(\delta; \theta; \hat{y}; y) \tag{2}$$

$$L = \alpha \ \ell_{ss} + 2\beta \ \ell_{cd} \tag{3}$$

where $\alpha$ refers to the weight parameter of segmentation loss functions, $\beta$ refers to the weight parameter of CDL, $y$ refers to the ground truth, $\hat{y}$ refers to the predict result, $\delta$ refers to the weight of changed samples in CDL, $\theta$ refers to the weight of unchanged samples in CDL.

The FL function considers both the imbalance of the sample sets and the difficulty level of the sample sets. This makes easily classified negatives comprise the majority of the loss and dominate the gradient, and the weight of the changed samples and unchanged samples computed by the FL function will be linear for samples of the same difficulty level. However, as we know, the imbalance of the sample sets and the difficulty level of the sample sets can both influence the loss function, so the relationship between the weights of the changed samples and unchanged samples cannot be linear.

$$CDL(\delta; \theta; \hat{y}; y) = \begin{cases} -(2 - \hat{y})^{\delta} \log(\hat{y}) & , y = 1 \\ -(1 + \hat{y})^{\theta} \log(1 - \hat{y}), y = 0 \end{cases} \tag{4}$$

The CDL function also contains two parameters, as shown in (4), which control the weights of the changed samples and unchanged samples, respectively, and consider the relationship between the weights of the changed samples and unchanged samples in a non-linear form. Through adjusting the relative values of the two parameters, we can solve problems of imbalanced sample sets, and the easily classified samples dominate the computed gradients.

## III. EXPERIMENTS

Many different datasets have been used for change detection, but the datasets generally only contain bitemporal images and their corresponding change labels. At present, only the public WHU building dataset [10] provides the semantic labels of the bitemporal images and the corresponding change labels. We therefore used this dataset to test the effectiveness of the proposed method.

### A. Dataset

The dataset covers an area where a 6.3-magnitude earthquake occurred in February 2011, in Christchurch, New Zealand. This dataset contains two scenes of images acquired at the same location in 2012 and 2016, with the semantic labels of the buildings and the change detection labels. Since the images are of 32507 ×15354 pixels, we divided both images into tiles of 256 ×256 pixels that did not overlap, and we finally formed the training sets, validation sets, and test sets with 6096, 762, and 762 images, respectively.

### B. Implementation Details

We used the mini-batch ADAM algorithm [10] to train the network. The batch size was set to 16 and the initial learning rate was set to 10-3. To measure the effect of the proposed network, we used the intersection over union (IoU), F1 score, recall, and accuracy as the metrics. For the comparison with other algorithms, data augmentation was achieved by a combination of random rotation and flipping of the images. For the loss function in the model training, the weight parameters of the two SSNs were set as 0.25, and the weight parameter of the change detection network was set as 0.5. We used PyTorch [11] as the deep learning framework. All the models were trained on a personal computer equipped with two NVIDIA GTX 1080 Ti GPUs.

### C. Ablation study

We evaluated the effect of each part of the network model on the WHU building dataset, as shown in Table I. In addition, we visualizethe results in Fig. 3. Based on an analysis in terms of performance and visual effect, it is confirmed that the proposed models and methods are effective. In addition to the methods we have proposed, data augmentation can be used to achieve a performance improvement of around 1.32% in the F1 score and 0.8% in the IoU, as shown in Table I

In order to improve the discriminative ability of the features extracted by the proposed model, we add the DAM to improve

TABLE II

Detailed Performance Comparison of the Proposed Network

| Method | Recall | Prec. | F1 | Iou |
|---|---|---|---|---|
| SCDN | 88.82 | 71.81 | 79.42 | 65.86 |
| SCDN +DAM | 87.00 | 79.22 | 82.93 | 70.84 |
| SCDN +DAM+FL | 82.36 | 84.63 | 83.48 | 71.65 |
| SCDN +DAM+CDL | 87.45 | 80.94 | 84.07 | 72.51 |
| SCDN +DAM+CDL +SSN | 89.63 | 88.29 | 88.95 | 80.12 |
| SCDN +DAM+CDL +SSN+DA | 89.35 | 90.15 | 89.75 | 81.40 |

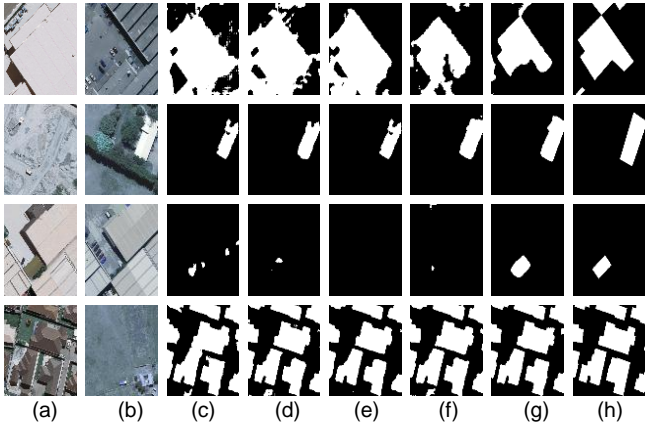

Fig. 3. Example results of the ablation study for the WHU building dataset. (a) 2016 images. (b) 2012 images. (c) Result of the Siamese change detection network (SCDN). (d) Result of SCDN with dual attention module (DAM). (e) Result of SCDN with DAM using focal loss. (f) Result of SCDN with DAM using change detection loss (CDL). (g) Result of SCDN with DAM using CDL. (h) Ground truth.

the dependence within channels and spatial positions through the self-attention mechanism, which improves the performance of the change detection. As shown in Table I, the F1 score is increased from 79.42% to 82.93%, and the IoU for the changed area is also increased from 65.86% to 70.84%.

In order to solve the problem of sample imbalance, we developed the CDL function by improving the FL function. It can be seen that using the FL function can improve the F1 score from 82.93% to 83.48% and the IoU from 70.84% to 71.65%. Using the CDL function achieves an F1 score of 84.07% and an IoU of 72.51%, which proves that the CDL function can obtain better results in the condition of sample imbalance, compared with the FL function.

In order to extract object-level discriminative features, we combine the change detection network and the SSNs, which achieves a significant improvement in all measures, as shown in Table I. Furthermore, adding the SSNs also achieves a better visual effect, in that the extracted changed area is complete and the boundaries are regular, as shown in Fig. 3(g).

### D. Performance Evaluation

We compared the experimental results of DTCDSCN obtained with the WHU building dataset with those of the other

TABLE I

Comparison of the Most Recent Change Detection Methods

| Method | Recall | Prec. | F1 | Iou |
|---|---|---|---|---|
| Improved-SegNet [6] | 78.28 | 88.23 | 82.96 | 70.88 |
| FC-EF [12] | 82.03 | 82.72 | 82.37 | 70.03 |
| FC-Siam-conc [12] | 74.46 | 73.68 | 74.07 | 58.82 |
| FC-Siam-diff [12] | 84.74 | 89.05 | 86.84 | 76.74 |
| Proposed DTCDSCN | **89.35** | **90.15** | **89.75** | **81.40** |

state-of-the-art change detection methods. During the test process, we used data augmentation. It took about three days for us to train our model, and if we had trained the models with 5-fold cross-validation, it would have taken us two weeks just to verify the proposed method, so we dropped the cross-validation for all the methods. The performance of the different change detection methods is reported in Table II. We also visually compare the proposed method and the improved SegNet method with the WHU building dataset in Fig. 1. Clearly, the proposed method can solve the problem of holes and irregular boundaries in the extracted changed areas relatively well by learning more discriminative object-level features.

The proposed method also solves the task of semantic segmentation, which is represented as building extraction for the WHU building dataset. We also analyzed the results of the building extraction, as shown in Table III, and the results of the change detection obtained using the post-classification comparison method based on the building extraction results, as shown in Table IV.

As Table III shows, the proposed method does not perform as well in semantic segmentation as UNet [13], with about a 5% lower IoU. We used the difference between the 2011 and

TABLE III

Comparison With Other Semantic Segmentation Methods

| Method | iou |
|---|---|
| U-Net | 93.06 |
| Proposed DTCDSCN | 88.86 |

2016 semantic segmentation labels and the change detection labels provided by the WHU building dataset as the ground truth to evaluate the change detection results based on the building extraction results, as shown in Table IV. We can see that the two results are unexpectedly different, which is worth analyzing. We believe that this is because of the inconsistency of the manual mapping of the bitemporal building labels and the deviation of the image registration, where the bitemporal building labels may have several pixels deviation at the building boundaries. This error is amplified when we subtract the semantic segmentation labels of the buildings, while the change detection labels provided in the WHU building dataset are subtracted by vector objects, which reduces the error accumulation. We also visualize the error in Fig. 4. Although the error has little impact on individual buildings, it has a greater impact on the overall visual effect and metrics due to fewer changes in the buildings. However, although a building extraction performance of 88.86% (IoU) has been achieved, the change detection achieves only 68% (IoU), which is poor compared with the proposed method. This proves that, for high-resolution pixel-level change detection, the method of post-classification comparison results in a large error without post-processing, and is far less effective than the direct classification method.

TABLE IV
EVALUATION OF THE POST-CLASSIFICATION COMPARISON METHOD

| Label | IoU |
|---|---|
| Subtracted labels | 60.87 |
| Dataset labels | 67.63 |

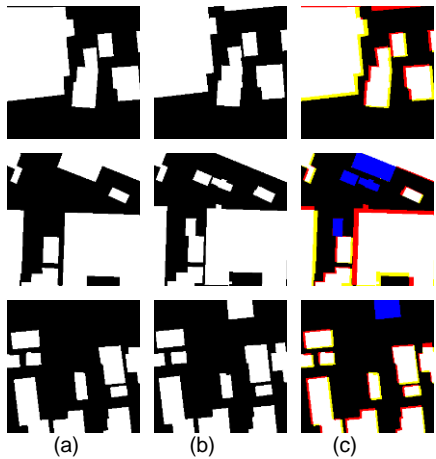

Fig. 4. Examples of the label error caused by manual mapping and image registration. (a) 2011 semantic segmentation labels. (b) 2016 semantic segmentation labels. (c) Comparison of the bitemporal labels. The yellow part is labeled as building in 2011 and non-building in 2016. The red part is labeled as non-building in 2011 and building in 2016. The blue part is the changed area.

## IV. CONCLUSION

In this letter, we have proposed a dual task constrained deep Siamese convolutional network (DTCDSCN) model for building change detection, which performs both semantic segmentation and change detection. This model contains a DAM that can be used to improve the feature representation by using the association between channels and spatial positions. In addition, the model is trained in an end-to-end manner using the CDL function to suppress the sample imbalance. The experimental results showed that the proposed method can achieve a better performance than the other state-of-the-art methods on the WHU building dataset. Our future work will include extending the model to general change detection and improving the method to an unsupervised or weakly supervised change detection method.